\documentclass[sigconf,nonacm]{acmart}
\setcopyright{acmlicensed}
\copyrightyear{2026}
\acmYear{2026}
\acmDOI{XXXXXXX.XXXXXXX}
\acmConference[Submitted to WebConf26]{Make sure to enter the correct
  conference title from your rights confirmation email}{}{}
\acmISBN{978-1-4503-XXXX-X/2018/06}
\usepackage{multirow}
\usepackage{booktabs}

\usepackage{xcolor}
\usepackage[most]{tcolorbox}
\tcbset{
  colback=gray!10,
  colframe=gray!50,
  boxrule=0.3pt,
  arc=1mm,
  left=1mm,
  right=1mm,
  top=0.5mm,
  bottom=0.5mm,
  fontupper=\small\ttfamily\color{black},
  enhanced,
  before skip=3pt,  
  after skip=2pt    
}

\renewcommand\footnotetextcopyrightpermission[1]{}
\begin{document}
\title{ToxiTwitch: Toward Emote-Aware Hybrid Moderation for Live Streaming Platforms}


\author{Baktash Ansari}
\authornote{Corresponding author.}
\email{baktash@uw.edu}
\affiliation{%
  \institution{University of Washington}
  \city{Bothell}
  \state{WA}
  \country{USA}
}


\author{Elias Martin}
\affiliation{%
  \institution{University of Washington}
  \city{Bothell}
  \state{WA}
  \country{USA}
}


\author{Afra Mashhadi}
\email{mashhadi@uw.edu}
\affiliation{%
  \institution{University of Washington}
  \city{Bothell}
  \state{WA}
  \country{USA}
}

\renewcommand{\shortauthors}{Ansari et al.}

\begin{abstract}
  
The rapid growth of live-streaming platforms such as Twitch has introduced complex challenges in moderating toxic behavior. Traditional moderation approaches, such as human annotation and keyword-based filtering, have demonstrated utility, but human moderators on Twitch constantly struggle to scale effectively in the fast-paced, high-volume, and context-rich chat environment of the platform while also facing harassment themselves. Recent advances in large language models (LLMs), such as DeepSeek-R1-Distill and Llama-3-8B-Instruct, offer new opportunities for toxicity detection, especially in understanding nuanced, multimodal communication involving emotes. In this work, we present an exploratory comparison of toxicity detection approaches tailored to Twitch. Our analysis reveals that incorporating emotes substantially improves the detection of toxic behavior. To this end, we introduce ToxiTwitch, a hybrid model that combines LLM-generated embeddings of text and emotes with traditional machine learning classifiers, including Random Forest and SVM. In our case study, the proposed hybrid approach reaches up to 80\% accuracy under channel‑specific training (with 13\% improvement over BERT and F1-score of 76\%). This work is an exploratory study intended to surface challenges and limits of emote‑aware toxicity detection on Twitch. 

\end{abstract}




\maketitle

Live-streaming platforms such as Twitch have become major arenas for real-time social interaction, where users express themselves through a mix of text, slang, and platform-specific visual symbols called emotes. While these emotes enrich communication, Twitch’s largely unmoderated chat environment has made both streamers and moderators vulnerable to toxic behaviors, including harassment, hate speech, and community-driven hostility~\cite{han2023hate,powell2023you}. Each streamer relies on volunteer moderators who manually review thousands of fast-paced comments~\cite{10.1145/3290605.3300390}, a form of invisible labor that carries emotional and psychological burdens~\cite{wohn2019volunteer,steiger2021psychological}. Twitch users collectively produce over 29 billion chat messages annually~\cite{electroiq2024twitchstats}, generating more than 80 GB of daily data—making comprehensive human moderation infeasible.
From a Responsible AI standpoint, moderation on Twitch poses distinctive challenges. Unlike text-only platforms, toxicity on Twitch is multimodal and culturally situated: meanings emerge not only from words but from visual symbols (emotes) that carry community-specific and evolving semantics~\cite{kim2022understanding,Moosavi2024}. The same emote may signify humor in one context and harassment in another, depending on the streamer, audience, and conversational flow. Automated systems that treat toxicity as static or universal risk misinterpreting cultural symbols, reinforcing biases, or over-censoring marginalized voices. Developing AI systems that are both context-sensitive and transparent about their limitations is therefore critical to ensuring safe and equitable participation in live-streaming communities.
Prior studies have begun to analyze Twitch toxicity using Natural Language Processing (NLP) methods. Kim et al.~\cite{kim2022understanding} demonstrated that emotes embedded within words can evade detection by standard text classifiers such as HateSonar~\cite{davidson2017automated}. Moosavi et al.~\cite{Moosavi2024} later showed that channel-specific emotes are disproportionately present in toxic interactions, yet assumed a fixed relationship between an emote and its toxicity label. These approaches, while valuable, do not address the contextual fluidity and cultural drift that define Twitch communication. Effective moderation requires models capable of reasoning about emotes’ meanings within their social and temporal contexts, and doing so responsibly, without overgeneralizing across communities.
Recent advances in Large Language Models (LLMs) and Vision–Language Models (VLMs)—such as GPT-4, Llama 3-8B-Instruct~\cite{Touvron2023LLaMAOA}, and Deepseek-R1~\cite{deepseekai2025deepseekr1incentivizingreasoningcapability}—offer new opportunities to capture this nuance by integrating textual and visual reasoning. Yet their performance in culturally embedded environments like Twitch remains underexplored. How well do these models interpret the social meaning of emotes? Can they generalize across communities with distinct linguistic norms? And how might their reasoning be augmented to mitigate bias and improve transparency?
This study investigates these questions through three research directions:
\begin{itemize}
\item \textbf{RQ1 (LLM Reasoning):} How well do reasoning-enabled LLMs perform in zero-shot toxicity classification across different Twitch communities?
\item \textbf{RQ2 (Emote-Enabled Prompting):} Can enriching prompts with contextual information about emotes enhance interpretability and reduce misclassification?
\item \textbf{RQ3 (Hybrid Model Efficacy):} Can combining LLM-generated embeddings with lightweight machine learning classifiers yield an efficient, interpretable, and latency-aware framework for real-time moderation?
\end{itemize}
To explore these questions, we focus on two distinct Twitch channels—HasanAbi, a “Just Chatting” channel where political discourse often sparks controversy, and LolTyler1, a gaming channel known for high-tempo interactions and gender-based toxicity. We collect 500 comments per channel and obtain human annotations from three coders, capturing generational and subjective perspectives on toxicity.
Our contributions are twofold:
\begin{itemize}
\item We investigate how contextual features extracted from emotes can enhance LLM reasoning for toxicity detection, showing that structured emote-aware cues reduce false positives by clarifying community-specific slang and symbolism.
\item We introduce ToxiTwitch, a hybrid architecture that integrates text and emote embeddings from LLMs with lightweight classifiers (Random Forest, SVM), achieving near 80\% accuracy with a 60 ms inference time per message—suitable for real-time deployment.
\end{itemize}

\noindent While our study is exploratory in scale, it provides insight into how multimodal reasoning and hybrid modeling can inform more responsible and context-aware AI moderation on dynamic social platforms.

\section{Related Work}\label{sec:related}

Toxicity on live streaming platforms like Twitch remains a growing concern as they expand beyond gaming~\cite{dreier2023toxicity}. Studies show topics like racism and politics attract more toxic comments than science or arts~\cite{10.1371/journal.pone.0228723,dreier2023toxicity}. The anonymity and pace of live chats reduce accountability, fueling toxic behavior through disinhibition~\cite{10.1371/journal.pone.0228723}, with consequences like mental health risks and hostile communities~\cite{hellopartner2022toxic}. Solutions include better moderation tools and community guidelines~\cite{dreier2023toxicity}. \citet{kim2022understanding} reverse-engineered toxic visual chats (e.g., emotes replacing letters), uncovering 196,448 new toxic comments via neural classifiers. \citet{dreier2023toxicity} found that male-hosted streams, larger audiences, and multiplayer/shooter games correlate with higher toxicity. To this end, we review previous research on detecting toxicity on Twitch and highlight the contributions of our paper based on the identified research gaps.

\subsection{Towards Automated Toxicity Detection on Live Streaming Platforms}

Twitch’s community guidelines allow streamers to appoint human moderators who manually filter chat messages, enforcing rules by blocking or suspending violators \citep{twitch2023guidelines}. However, moderators face significant challenges, including processing large volumes of messages in real-time and enduring harassment while mediating conflicts \citep{10.1145/3359157,10.1145/3290605.3300390,10.1145/3567568}. To alleviate these burdens, researchers have proposed integrating automated moderation tools to assist human moderators in maintaining a less toxic environment. \citet{Moon2023AnalyzingNV} focuses on detecting norm violations, including toxic language, in live-streaming platforms like Twitch. It uses automated methods, specifically training models with contextual information, to improve moderation performance. They conclude that models trained with existing data sets perform poorly in detecting toxic messages and motivate the development of specialized approaches for Twitch. Recent research has explored various computational approaches to detect and mitigate toxicity in live streaming platforms. We organize prior work into three key themes: (1) toxicity detection methods, and (2) platform-specific challenges. Efforts to improve real-time detection include model efficiency and cross-platform adaptation. \citet{oikawa2022stacking} proposes a stacking-based method for Japanese live chats, prioritizing inference speed for offensive language. Similarly, \citet{gao2020offensive} leverages transfer learning from Twitter to detect offensive content on Twitch, addressing data scarcity. However, most of these solutions lack the incorporation of Twitch-specific features. We take inspiration from~\cite{Moosavi2024}, which uses toxicity detection on Twitch by creating an emote embedding space. They use a RoBERTa-based toxicity classifier to identify seed emotes and then discover other toxic emotes through embedding association. The paper also discusses the limitations of NLP models in detecting toxicity in Twitch chat and proposes an active learning moderating system.

A core limitation is the difficulty in interpreting complex linguistic features such as sarcasm, metaphors, plus the addition of platform-specific emotes, often leading to both false positives and missed detections \cite{hu2024toxicity,nobata2016abusive}. Even sophisticated tools like the PERSPECTIVE API struggle with interpretability, as their toxicity scores reflect prediction confidence rather than severity, limiting their practical utility \cite{welbl2021challenges}. Additionally, there is an over-reliance on lexicon-based methods and models trained on data from other platforms, which fail to capture Twitch's unique contextual dynamics \cite{dreier2023toxicity,poyane2018toxic}. While lexicon-based approaches can identify known toxic patterns, they are ineffective against novel or evolving forms of toxicity~\cite{ali2025evolving}. The scarcity of labeled data poses another major constraint, particularly for token-level toxicity detection, where transformer models require extensive fine-tuning datasets that are often unavailable \cite{jain2021token}. Current methods also grapple with false positives and negatives, especially in real-world scenarios where toxic instances are rare. Even low false positive rates become problematic at scale, generating excessive false alarms \cite{hu2024toxicity}.

While LLMs show promise for toxicity detection, they introduce new challenges. These include sensitivity to prompt quality, high computational costs, and latency issues when processing large volumes of text \cite{njeh2025llm}. Advanced models like ToxBuster acknowledge they cannot fully replace human moderators and are unsuitable for fully automated moderation \cite{yang2023toxbuster}. These limitations demonstrate that no single automated solution can perfectly address toxicity detection, especially in dynamic environments like live streaming platforms. This underscores the necessity for hybrid approaches and continuous methodological improvements \cite{aldahoul2024hybrid,zhu2024advances}.

\subsection{Emotes and Toxicity Detection Challenges}

Emotes in Twitch have become an integral part of online communication and research has shown that they play a significant role in toxic communication  in more nuanced and covert ways, making them  a crucial clue for comprehensive toxicity detection systems~\cite{kim2022understanding,Moosavi2024}. The visual nature of emotes allows users to convey complex emotions and intentions that may \textit{not} be easily captured by traditional text-based methods. In this vein, studies have demonstrated that certain emotes are frequently associated with toxic behavior, and their usage patterns can provide valuable insights into the overall toxicity of a conversation~\cite{kim2022understanding}. However, the context-dependent nature of emotes adds complexity to toxicity detection, as the same emote may have different connotations depending on the surrounding text and conversation~\cite{davange2024toxic}. To address this challenge, \citet{Moosavi2024} mapped an embedding space of channel and global emotes, which could be used to discover similarity of emotes across communities and thus aid with knowledge transfer of toxicity detection models across communities.  ~\citet{kim2022understanding}  developed labeled datasets and neural network classifiers specifically designed to identify  toxic chats that would be missed by traditional methods. They demonstrated that analyzing emotes in chat conversations can catch an additional 1.3\% of toxic instances out of 15 million utterances on Twitch. This finding underscores the need for more sophisticated detection methods, as traditional approaches may miss visual forms of toxicity.
While prior work such as \citet{kim2022understanding} and \citet{Moosavi2024} provide important insights into emote usage, their experimental setups differ in scope and dataset composition, making direct comparison infeasible. Instead, we situate our work,  ToxiTwitch, as complementary: our contribution lies in demonstrating the latency–accuracy tradeoff of hybrid architectures.




\section{Dataset and Annotation}\label{sec:data}
This section outlines our data collection methods and the human annotation technique. 

\begin{figure*}[ht]
\centering
    \includegraphics[width=0.48\textwidth]{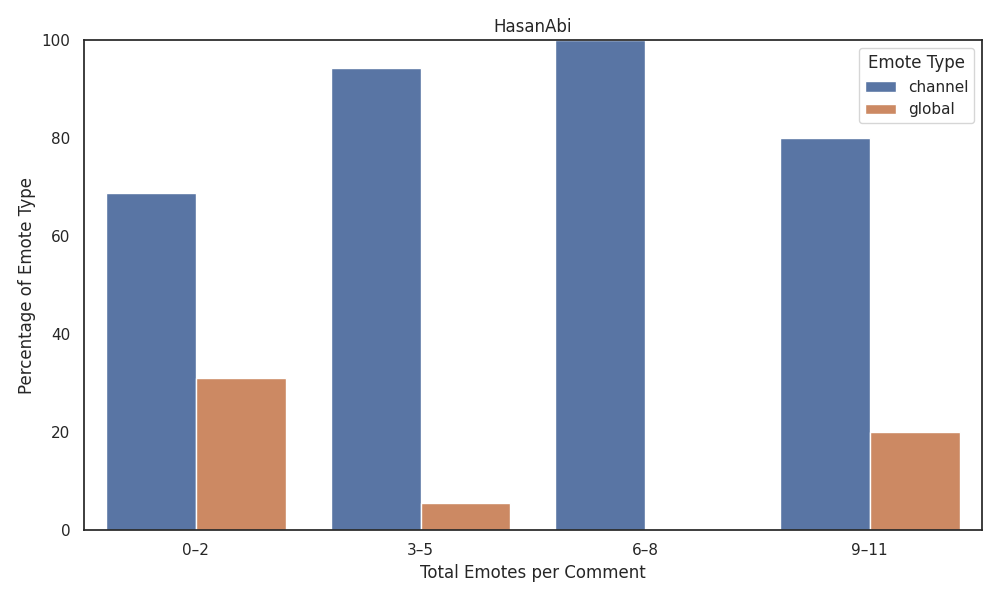}
  \includegraphics[width=0.48\textwidth]{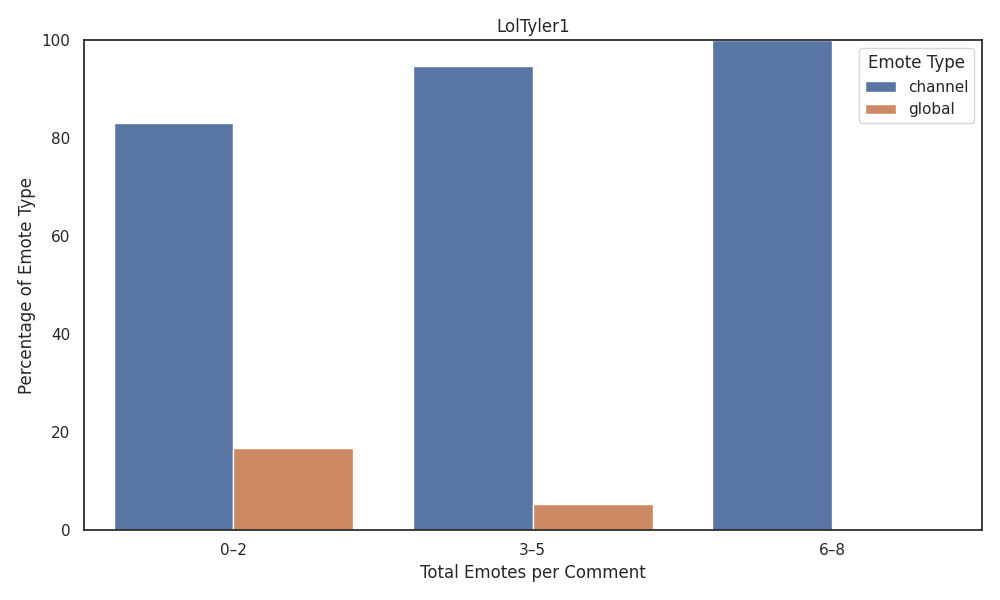}
  \caption{Percentage of Global vs. Channel Emotes by Number of Emotes per Comment}
  \label{fig:emote_percentage}
\end{figure*}

\subsection{Twitch Comments}
In curating a dataset and selecting Twitch streams suitable for this task, we identified two topics of just-chatting and gaming. Within channels related to these topics, we then selected two channels that are consistently known as the top 10 popular but also controversial channels on Twitch. In the gaming category, this resulted in a channel dedicated to the League Of Legends game named \textit{LolTyler1}\footnote{https://www.twitch.tv/loltyler1}. In just chatting category, we selected the \textit{HasanAbi}\footnote{https://www.twitch.tv/hasanabi} channel, which is known for its coverage of controversial discussions. 

For HasanAbi, data was recorded for November 17th 2022, and captured the entirety of the channel's stream chat logs. Similarly, for LolTyler1, data was captured during one of his streams on November 11, 2022.  Overall, we captured 500 comments from each of the two channels.

These comments are in length typically 10-15 words for HasanAbi and 0-5 words for LolTyler1, indicating that the majority of comments on Twitch are relatively short.
This makes content moderation a unique challenge as the majority of comments lack the length and context that more traditional forms of moderation might be trained to perform on platforms with lengthier comments, such as Twitter or Wikipedia.


Each HasanAbi comment contains on average 1.40 emotes, showing a noticeable use of emotes in communication. Among the emotes used, a large majority are channel-specific emotes, making up over 80\% of all emote usage across all emote count ranges. The rest are global emotes, which appear more often in comments with fewer total emotes. This shows that viewers in HasanAbi’s stream mostly use emotes that are unique to the channel, which usually require a subscription to unlock.

Similarly, the LolTyler1 dataset also shows emote usage but at a lower rate. On average, each comment includes 0.72 emotes. Like HasanAbi’s data, most of the emotes used are channel-specific, with over 80\% of all emotes being tied to the channel. As the total number of emotes in a comment increases, the share of channel emotes also increases, reaching nearly 100\% in comments with more than 5 emotes.

Figure~\ref{fig:emote_percentage} shows the percentage of global and channel-specific emotes used in Twitch comments, illustrating patterns in emote popularity. These results show the strong presence of emote culture in Twitch chats, especially through the use of channel-specific emotes. This makes it harder for NLP models to understand the content, since many of the most common emotes are not available outside the specific channel and often carry unique meanings within that community.

\subsection{LLM's Knowledge of Emotes}\label{sec:llmsknow}

Given the popular nature of global emotes and their widespread presence in public data, it is highly likely that these LLMs encountered them during training, enabling them to incorporate knowledge of these emotes into their language models. We investigate the extent to which LLMs have acquired knowledge of Twitch's global and channel emotes, focusing on their ability to interpret and explain the meaning of an emote.

\noindent{\textbf{Global Emotes:}} To test LLM's knowledge of the global emotes, we randomly selected 100 global emotes and performed the following analysis.  We prompted Llama-3-8B to explain the meaning of each emote. We then manually inspected its answers. Out of 100 global emotes Llama described 93 of those emotes correctly.  Through our analysis we found that LLMs could accurately interpret global emotes in both isolated and contextually rich environments, simulating real-world Twitch conversations. We observed similar results with Deepseek. 

\noindent{\textbf{Channel-specific Emotes:}} As channel-specific emotes are smaller sets, we ran the entire set of HasanAbi channel emotes and Loltyler emotes and prompted the LLMs whether they knew these emotes. Out of the 318 Loltyler channel-emotes Llama-3-8B could describe only 72 emotes (23\%). Out of 736 HasanAbi channel-emotes, Llama-3-8B only could provide description of 145 emotes (20\%).

For Deepseek, we observe that the model tries to convince itself through reasoning steps that it knows the emote. It often relies on contextual clues, such as the emote text and the name of the channel, to generate an explanation. The final answer is usually given with a high degree of uncertainty, so it is unclear whether the model truly understands the emote or is just hallucinating a response.

\subsection{Ground-Truth Human Annotation}

\subsubsection{Annotation Guideline}
In order to assess the performance of various models, we require a human-annotated validation set.  Toxicity labeling is a challenging annotation task due to the inherent subjectivity and ambiguity involved. To address this, we developed a detailed annotation code-book \footnote{\url{https://anonymous.4open.science/r/toxitwitch-C5A7/codebook.pdf}} and recruited three annotators. While all annotators currently reside in the United States, they brought diverse cultural perspectives. Annotators were asked to classify comments as toxic or non-toxic, and for toxic comments, to assign one or more relevant toxicity categories (e.g., obscene, threat, insult, identity attack, sexually explicit)  as in ~\cite{jigsaw-toxic-comment-classification-challenge}. 
Annotators were told to look up the meanings of emotes to assist in their decisions.  To this end and to minimize misinterpretations, annotators consulted emote dictionaries or platforms such as BTTV and FFZ, particularly for channel-specific emotes that may not be widely recognized outside certain Twitch communities. This additional step was critical to ensure the correct assessment.

\subsubsection{Annotated Dataset}
In order to balance the dataset and to increase the chances of toxic comments in the annotation pool, we first applied a toxicity detection model to our datasets described above.  We chose a DistilBERT model trained on ToxiGen Dataset~\cite{hartvigsen2022toxigenlargescalemachinegenerateddataset} to perform the initial pass and classify the comments into toxic/non-toxic. ToxiGen~\cite{hartvigsen2022toxigenlargescalemachinegenerateddataset} is a toxicity dataset generated using an LLM to analyze and address toxicity in language. It is designed to capture diverse forms of toxic content, often organized into categories representing different types of harmful speech. ToxiGen stands out due to its focus on the subtle, contextual, and diverse aspects of toxic language.

Out of the comments that were predicted as toxic by Distilbert-Toxigen, we selected 500 comments that higher prevalence of emotes in them for each of channels. All three annotators labeled each data point and the label with the majority votes was chosen. 

\section{Experimental Setting}\label{sec:method}



\subsection{Large Language Models}
LLMs have emerged as powerful tools for toxicity detection, leveraging their deep contextual understanding and vast knowledge to identify harmful content across multiple languages and domains. We leverage two open-source implementations of LLMs, namely LLaMA 3-8B-Instruct~\cite {Touvron2023LLaMAOA} and Deepseek-R1-Distill
~\cite{deepseekai2025deepseekr1incentivizingreasoningcapability}, for our analysis. Both models are comparable in scale, with LLaMA~3-8 B-Instruct and Deepseek-R1-Distill containing approximately 8 billion parameters each. For all our experiments, we used hyperparameters temperature = 0.5 and a probability threshold for sampling tokens (top\_p) = 0.9 for consistency.

\subsubsection{LLaMA 3-8B-Instruct~\cite{Touvron2023LLaMAOA}:}
Llama-based models, developed by Meta, are highly effective for toxicity detection due to their advanced language understanding and training on large, diverse datasets~\cite{nguyen2023finetuningllama2large,luong-etal-2024-realistic}.
These models can identify harmful language patterns, including hate speech and cyberbullying, even in subtle contexts. Compared to traditional models like RoBERTa, Llama's generative capabilities make it adaptable to evolving toxic behaviors, particularly in real-time environments like social media~\cite{luong-etal-2024-realistic}.

\subsubsection{Deepseek-R1~\cite{deepseekai2025deepseekr1incentivizingreasoningcapability}:}
We chose Deepseek as the second model for comparison in our study. DeepSeek-R1-Distill is optimized for high-throughput, low-latency environments where both speed and precision are essential for real-time applications like content moderation. While larger models (e.g., GPT-4, Claude) achieve state-of-the-art results on broad NLP benchmarks, DeepSeek-R1-Distill strikes a practical balance between model efficiency, inference cost, and task-specific accuracy, while remaining open-weight and deployable on commodity hardware. Its design prioritizes scalability for live platforms, making it a compelling candidate for comparison against both heavyweight LLMs (e.g., LLaMA 3-8B-Instruct) and compact task-specific models (e.g., DistilBERT). By evaluating DeepSeek-R1-Distill alongside these alternatives, we assess for real-world viability, particularly in scenarios demanding low-resource operation without sacrificing robustness. 

\subsection{Prompt Baselines}\label{sec:prompts}

Prompts are text fragments passed into LLMs. In the past years, research has shown that designing prompts is notoriously challenging, and even small differences in the prompt template can significantly impact the answer of the LLMs~\cite{liu2023pre}. 
In this work, we focus on soft prompting, that is the content of the comment is integrated into the prompt. We also focus on \textit{zero-shot} prompting, that is, the prompt given to the LLM does not include any examples or demonstration~\cite{kojima2022large}. 

  There are several approaches for zero-shot prompting pre-trained LLMs that have been shown to help LLMs in implementing step-by-step reasoning. One such approach is Chain-of-thought (CoT)~\cite{10.5555/3600270.3602070}. CoT is a robust method for reasoning, particularly effective in addressing open-ended questions. It works by breaking down complex problems into a series of logical, coherent steps, facilitating clearer and more structured reasoning. Previous research on LLMs' reasoning on social media data has shown that COT outperforms vanilla prompting by far~\cite{amirizaniani2024llms,Ho2022LargeLM,Wang2022TowardsUC}. By employing COT prompting, we believe that the LLMs would be able to reason on the \textit{global} emotes embedded in comments and detect toxicity more accurately. The COT prompt that we adapted to this study is as follows:

\begin{tcolorbox}
``If you were a moderator on the Twitch streaming platform, imagine encountering a comment with the following content. Your objective is to flag the comment as <toxic> or <non-toxic> based on the following toxicity definitions: <obscene, threat, insult, identity attack, sexually explicit>. Consider this reasoning step by step.
Return a JSON object with four fields: ``Is it toxic,'' which is either ``yes'' or ``no'' depending on whether you would remove this comment from the stream; which toxicity category it falls into; and “explanation,” which provides a reason for your decision; and whether your explanation included any emotes present in the message and whether the emote conveyed toxicity.''
\end{tcolorbox}

\section{Our Proposed Method}
We propose  a modular, emote-aware AI moderation agent designed to operate in real-time, high-volume social platforms such as Twitch. Unlike static classifiers, our model, ToxiTwitch, functions as a contextually grounded reasoning agent that continuously interprets user-generated chat streams through three tasks:

\textbf{Perceiving} multimodal inputs (text and emotes) through lightweight preprocessing and embedding generation;

\textbf{Reasoning} about the contextual intent of emotes using prompt-augmented LLMs, which act as the agent’s inference engine;

\textbf{Acting} via a decision-making module—realized through traditional classifiers—that flags  toxic content with low latency and high precision.

We describe perceiving and reasoning task as part of a combined stage referred to as Emote-Enabled-Prompting, and the acting task of the agent in the ToxiTwitch model.

\subsection{Emote-Enabled-Prompting}\label{emote-handling}

Previous research has shown that providing extra context as part of prompt can enhance LLM's social reasoning abilities~\cite{amirizaniani2024llms}. To this end, we  incorporate additional information related to channel emotes into the LLMs in two ways: 

\begin{figure}[ht]
\centering

  \includegraphics[width=0.5\textwidth]{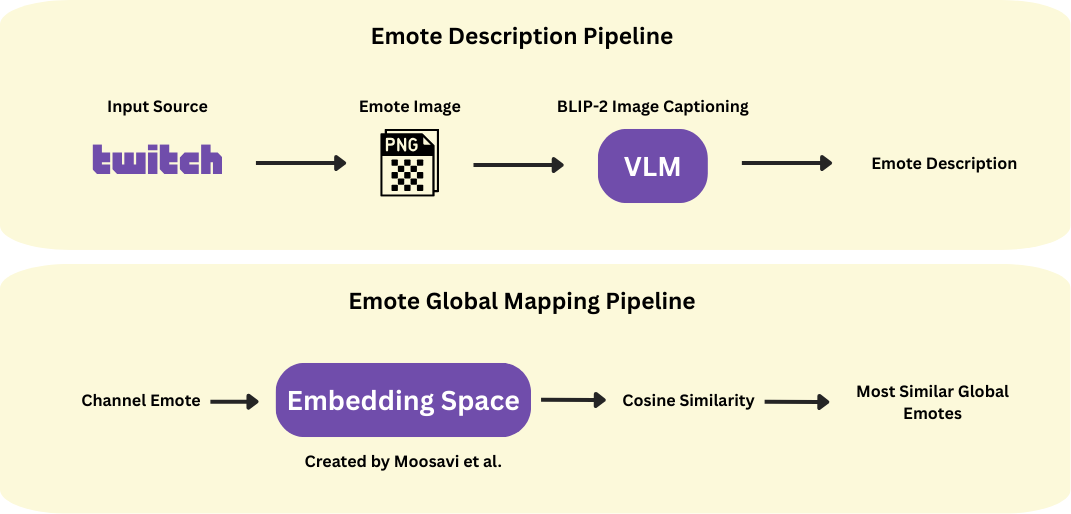}
  \caption{Emote Description Generation and Global Mapping Pipelines}
    \label{fig:pipelines}
\end{figure}

\subsubsection{{Emote-Description (ED):}} 
For each channel emote, we construct a description of the emote. The system relies on a pipeline that maps each channel-specific emote to a description of what the emote represents using multimedia processing. Figure~\ref{fig:pipelines} illustrates our process for generating these emote descriptions. We used the BLIP-2 model~\cite{li2023blip2bootstrappinglanguageimagepretraining} to generate descriptions for each emote image. Finally, we created a dictionary for each emote along with its description. We modify the COT prompt template and append the emote name followed by its description for every emote that was seen in the comment:

\begin{tcolorbox}
``Consider that <channel-emote> in this comment is described as <emote description>. 
Perform step-by-step reasoning.''
\end{tcolorbox}

\subsubsection{Emote-Global-Mapping (EGM):}
Alternatively and given that LLMs have a good knowledge of global emotes (as shown in Section 3.3), we find the top most similar global emotes for each channel emotes. To do this, we used the embedding space created by~\citet{Moosavi2024} which was based on the top 2379 English-speaking channels in  2022. Unlike other available Twitch embedding spaces, this dataset includes mapping of global and, importantly \textit{channel emotes},  enabling us to calculate and retrieve the similarity of emotes across different channels and communities. As the embedding space was created using Word-2-Vec, we applied cosine similarity to measure relations between channel to global emotes. 
To provide LLMs with contextual information about the \textit{channel} emotes, we use this embedding space   to map the top three closest \textbf{global} emotes to any given channel emote. 
Figure~\ref{fig:pipelines} illustrates our process for extracting the most similar global emotes to our channel emote. We \textit{append} the extracted  top global emotes to the zero-shot CoT prompt template: 

\begin{tcolorbox}
``Consider that <channel-emote> in this comment is closest to Global Emotes:(GE1,GE2,GE3). Perform step-by-step reasoning.''
\end{tcolorbox}
 

\subsection{ToxiTwitch Model}

Our framework processes Twitch comments in two stages: (1)~contextual embedding extraction using Large Language Models (LLMs), and (2)~toxicity classification via simple machine learning models. The pipeline is designed to handle the unique characteristics of platform-specific emotes, fast-paced interactions, and contextual dependencies.

\subsubsection{Embedding Generation}

To generate semantically meaningful representations of chat messages, we leverage LLMs to obtain dense embeddings. Given a sequence of chat messages 
\( X = \{x_1, x_2, \dots, x_n\} \), 
where each \( x_i \) denotes a single user message, our objective is to encode each message into a fixed-size vector capturing both lexical and contextual semantics.
The tokenized input is then passed through the LlaMA transformer encoder, which produces token-level contextual embeddings. For LLaMA 3-8B-Instruct, the embedding for message \( x_i \) is:

\begin{equation}
H_i = \text{Llama}(x_i) \in \mathbb{R}^{L \times d}
\label{eq:token_embeddings}
\end{equation}

where \( L \) is the number of tokens in the message after tokenization, and \( d = 4096 \) is the hidden dimensionality of the model. The output matrix 
\( H_i = [h_{i1}, h_{i2}, \dots, h_{iL}] \) 
consists of token embeddings \( h_{ij} \in \mathbb{R}^d \) for each token \( j \) in message \( i \). We extract these embeddings from the final layer of the transformer to capture the highest level of semantic abstraction.

To obtain a message-level representation, we apply mean pooling over all token embeddings:
\begin{equation}
e_i = \frac{1}{L} \sum_{j=1}^{L} h_{ij}, \quad e_i \in \mathbb{R}^{d}
\label{eq:mean_pooling}
\end{equation}
This operation yields a single vector \( e_i \) representing the entire message content. For Deepseek-R1, we follow an analogous procedure but with \( d = 2048 \). These message-level embeddings \( \{e_1, \dots, e_n\} \) serve as input features for the downstream task of toxicity classification. To account for the contextual nature of conversational toxicity, we also explored incorporating emotes into the text embeddings.

\subsubsection{Emote Handling Strategies}

 A critical challenge in processing Twitch chat is the prevalence of platform-specific emotes that carry semantic meaning but may be tokenized sub-optimally by general-purpose language models. We leverages the same two distinct strategies for emote handling as mentioned in Section~\ref{emote-handling}. 
For each strategy, we compute separate embedding sets and evaluate their downstream classification performance.

\subsubsection{Toxicity Classification}

The generated embeddings serve as feature vectors for toxicity classification models. We use traditional machine learning models, Random Forest~\cite{breiman2001random} and Linear SVMs~\cite{cortes1995support} for the classification stage due to their efficiency and simplicity in compared to LLMs. We hypothesize that by using simple traditional machine learning models, we can achieve performance comparable to zero-shot experiments with the added advantage of lower computational cost. In subsequent sections, we evaluate the effectiveness of multiple embedding strategies, raw text embeddings, emote description augmentation, and emote replacement across different classification architectures. We use repeated stratified cross‑validation with separate held‑out test sets per fold; we retain the original description to reflect the exploratory setup.



\section{Results}
In this section, we present the results of our study in addressing each of the research questions.

\subsection{Baseline Performance of LLMs}
We first compare the models in terms of their performance to understand how well they perform when classifying toxicity on Twitch in different communities. Table~\ref{tab:model-performance-1} presents the results of this comparison. As can be seen, both LLMs perform comparatively to each other and across two different communities. Although the models exhibit high recall, they suffer in terms of low precision, that is, suffering from a high number of false positives. These results confirm our hypothesis that LLMs lack precision and due to their reasoning become over sensitive in flagging toxicity when they lack  information about community nuances.

\begin{table}[h!]
\centering
\small
\caption{Model Performance Comparison Across Datasets}
\begin{tabular}{llcccc}
\hline
\textbf{Dataset} & \textbf{Model} & \textbf{Acc.} & \textbf{Precision} & \textbf{Recall} & \textbf{F1} \\
\hline
   & Llama      & 0.84 & 0.30 & 0.81 & 0.39 \\
 HasanAbi  & DeepSeek       & 0.78 & 0.25 & 0.90 & 0.39 \\
\hline
   & Llama      & 0.63 & 0.37 & 0.90 & 0.52 \\
LolTyler1   & DeepSeek       & 0.53 & 0.32 & 0.87 & 0.47 \\
\hline
\end{tabular}
\label{tab:model-performance-1}
\end{table}

\subsection{Impact of Emote-Enabled Prompting}
Table~\ref{tab:model-performance-appended} presents the result of our emote-enabled prompting technique for both ED and EGM across models and communities. As we can see by including information on emotes, in all cases the models improve in their reasoning about toxicity in terms of F1 Score. While we do not observe a drastic increase in precision, we can see that in some cases such as Llama-ED on Hasanabi, the description of the channel emotes has guided the model to achieve higher recall, while maintaining  precision. It is worth noting that as we described in Figure~\ref{fig:emote_percentage} there is a relatively low usage of emotes in our sample, which explains the lack of significant changes in the emote-enabled prompting compared to the baseline.

\begin{table}[h!]
\centering
\small
\caption{Emote-Enabled-Prompting Performance Comparison Across Datasets. Underlined values denote the highest performance for the channel, and stars denote p-value with significant improvement of F1 score in comparison to baseline results of Table 1. p-values $<$ 0.05 are indicated with ***. }
\begin{tabular}{llcccl}
\hline
\textbf{Dataset} & \textbf{Model} & \textbf{Acc.} & \textbf{Precision} & \textbf{Recall} & \textbf{F1} \\
\hline
   & Llama-ED       & \underline{0.84} & \underline{0.31} &  \underline{0.87} &  \underline{0.46} *** \\
HasanAbi   & Llama-EGM      & 0.83 & 0.30 & 0.84 & 0.44  *** \\
   & DeepSeek-ED    & 0.79 & 0.26 & 0.88 & 0.45 *** \\
   & DeepSeek-EGM   & 0.77 & 0.25 & 0.84 & 0.38   \\
\hline
   & Llama-ED       & 0.64 & 0.38 & 0.85 & 0.53 *** \\
LolTyler1   & Llama-EGM      & \underline{0.61} & \underline{0.37} & \underline{0.92 }& \underline{0.53} *** \\
   & DeepSeek-ED    & 0.63 & 0.36 & 0.80 & 0.50 *** \\
   & DeepSeek-EGM   & 0.62 & 0.36 & 0.85 & 0.51 *** \\
\hline
\end{tabular}
\label{tab:model-performance-appended}
\end{table}

\subsection{Evaluating ToxiTwitch: Hybrid Model for Toxicity Detection}



\subsubsection{ToxiTwitch Performance}
We present the performance of various embedding strategies on two streamers—HasanAbi and LolTyler—using Random Forest and Linear SVM classifiers in Tables~\ref{tab:performance} and~\ref{tab:performance-lol} respectively. Our findings reveal that the addition of EGM significantly outperforms all other embedding configurations. For HasanAbi, Textual Embedding + EGM achieves the highest F1 score of 0.79 (RF) and 0.79 (SVM), along with substantial gains in accuracy (0.86 and 0.83, respectively). Similarly, Textual Embedding + EGM attains the strongest performance for LolTyler1, with F1 scores of 0.78 (RF) and 0.78 (SVM). Providing emote descriptions also enhances the base embeddings. When ED is added to Llama or Deepseek embeddings, performance improves consistently across all metrics. For instance, in the HasanAbi dataset. However, the magnitude of improvement from ED is less than that of EGM, indicating that replacing channel emotes with global emotes offers richer context than descriptions (ED) alone. We also find that Llama-based embeddings generally outperform Deepseek-based ones across both datasets. For example, Textual Embedding + EGM achieves an F1 of 0.79 (RF) for HasanAbi, whereas Textual Embedding + EGM scores 0.75. This trend is consistent in the LolTyler1 dataset, indicating that Llama embeddings provide more semantically rich representations for downstream classification tasks in this domain. Lastly, Random Forest slightly outperforms Linear SVM in most configurations, particularly in Recall and Accuracy, suggesting that ensemble-based models may be better suited to capture the subtle variations in toxic content.

\begin{table*}[ht]
\centering
\small
\caption{Performance comparison of embedding techniques with highest values of each metric being underlined. (HasanAbi)}
\label{tab:performance}
\begin{tabular}{lllllllll}
\toprule
\multirow{2}{*}{Embedding Technique} & \multicolumn{4}{c}{Random Forest} & \multicolumn{4}{c}{Linear SVM} \\
\cmidrule(lr){2-5} \cmidrule(lr){6-9}
 & Precision & Recall & F1 & Accuracy & Precision & Recall & F1 & Accuracy \\
\midrule
Llama Text Embedding       & 0.64 & 0.79 & 0.71 & 0.75 & 0.61 & 0.74 & 0.67 & 0.70 \\
Llama Text + ED            & 0.68 & 0.81 & 0.74 & 0.76 & 0.68 & 0.79 & 0.73 & 0.74 \\
Llama Text + EGM           & \underline{0.74} & \underline{0.86} & \underline{0.79} & \underline{0.86} & \underline{0.76} & \underline{0.83} & \underline{0.79} & \underline{0.83} \\
Deepseek Text Embedding    & 0.62 & 0.70 & 0.66 & 0.77 & 0.61 & 0.70 & 0.65 & 0.77 \\
Deepseek Text + ED         & 0.70 & 0.75 & 0.72 & 0.78 & 0.70 & 0.72 & 0.71 & 0.72 \\
Deepseek Text + EGM        & 0.71 & 0.80 & 0.75 & 0.78 & 0.73 & 0.76 & 0.75 & 0.78 \\
\bottomrule
\end{tabular}
\end{table*}

\begin{table*}[ht]
\centering
\small
\caption{Performance comparison of embedding techniques with highest values of each metric being underlined (LolTyler1)}
\label{tab:performance-lol}
\begin{tabular}{lllllllll}
\toprule
\multirow{2}{*}{Embedding Technique} & \multicolumn{4}{c}{Random Forest} & \multicolumn{4}{c}{Linear SVM} \\
\cmidrule(lr){2-5} \cmidrule(lr){6-9}
 & Precision & Recall & F1 & Accuracy & Precision & Recall & F1 & Accuracy \\
\midrule
Llama Text Embedding       & 0.61 & 0.80 & 0.69 & 0.71 & 0.58 & 0.75 & 0.66 & 0.67 \\
Llama Text + ED            & 0.65 & 0.82 & 0.73 & 0.73 & 0.65 & 0.80 & 0.72 & 0.73 \\
Llama Text + EGM           & \underline{0.70} & \underline{0.87} & \underline{0.78} & \underline{0.79} & \underline{0.72} & \underline{0.84} & \underline{0.78} & \underline{0.78} \\
Deepseek Text Embedding    & 0.59 & 0.71 & 0.64 & 0.65 & 0.58 & 0.72 & 0.64 & 0.65 \\
Deepseek Text + ED         & 0.66 & 0.76 & 0.71 & 0.71 & 0.67 & 0.73 & 0.70 & 0.70 \\
Deepseek Text + EGM        & 0.67 & 0.82 & 0.74 & 0.74 & 0.70 & 0.77 & 0.73 & 0.73 \\
\bottomrule
\end{tabular}
\end{table*}


\subsubsection{Benchmark Comparison}
To assess the efficacy of our proposed hybrid model for toxicity detection on Twitch, we perform a comparative evaluation against three publicly available transformer-based baselines: Detoxify \cite{unitaryai_detoxify}, HateSonar \cite{hateSonar}, and DistilBERT-ToxiGEN \cite{hartvigsen-etal-2022-toxigen} and our model ToxiTwitch. For ToxiTwitch, we selected the best performing setting based on previous observation (Section 6.3.1) which corresponds to Random Forest on  Llama Text + EGM. To this end, we collected two \textbf{different} days from our original training data of HasanAbi and LolTyler1 that were previously not seen by ToxiTwitch. For each streamer, we inferred toxicity labels in real time until 200 toxic and 200 non-toxic messages were identified by ToxiTwitch. We then manually annotated all 800 messages (400 per streamer) to establish human ground truth labels following the same procedure stated in Section 3.3.2.


In addition, we measured inference latency (in milliseconds) on a single-core CPU Thermal Design Power (TDP) of   29 Watts and using PyTorch, simulating a constrained but realistic deployment setting for live moderation systems. Our latency metric reflects the time required to process a single average-length Twitch message (16–24 tokens).

ToxiTwitch achieves competitive performance   in terms of F1-score and precision, demonstrating its ability to reduce false positives while retaining high recall (Tables~\ref{tab:Performance_hA}) while demonstrating the viability of hybrid emote-aware models. Notably, our model achieves an F1 of 0.73 and 0.71 on HasanAbi and LolTyler datasets, respectively, compared to 0.68/0.65 for HateSonar and substantially lower scores for Detoxify and ToxiGEN. These improvements can be attributed to the hybrid architecture’s capacity to leverage both linguistic semantics and emote-based contextual cues common in Twitch discourse. While HateSonar achieves slightly higher recall, its precision is lower, suggesting more false positive detections.

Regarding latency, ToxiTwitch remains competitive, with an average inference time of 60 ms per message—within the threshold for near-real-time moderation and only marginally higher than Detoxify (45 ms), which is optimized for speed but shows reduced classification accuracy.

\begin{table}[h!]
\centering
\caption{Benchmark Comparison of SOTA models. The underline shows the highest performance metric/ lowest latency. }
\label{tab:Performance_hA}
\small
\resizebox{\columnwidth}{!}{
\begin{tabular}{l|c|c|c|c|c}

\textbf{Model} & \textbf{Prec.} & \textbf{Rec.} & \textbf{F1} & \textbf{Acc.} & \textbf{Latency (ms)} \\ \hline
\multicolumn{6}{c}{\textbf{HasanAbi}} \\ 
\hline
Detoxify~\cite{unitaryai_detoxify} & 0.45 & 0.69 & 0.54 & 0.67 & \underline{45} \\ \hline
HateSonar~\cite{hateSonar} & 0.55 & 0.88 & 0.68 & 0.77 & 70 \\ \hline
DistilBERT-ToxiGEN~\cite{hartvigsen-etal-2022-toxigen} & 0.24 & 0.72 & 0.36 & 0.43 & 55 \\ \hline
\textbf{ToxiTwitch (ours)} & \underline{0.63} & 0.87 & \underline{0.73} & \underline{0.80} & {60} \\ \hline

\multicolumn{6}{c}{\textbf{LolTyler1}} \\ 
\hline
Detoxify~\cite{unitaryai_detoxify} & 0.38 & 0.75 & 0.50 & 0.63 & \underline{45} \\ \hline
HateSonar~\cite{hateSonar} & 0.47 & 0.93 & 0.62 & 0.72 & 70 \\ \hline
DistilBERT-ToxiGEN~\cite{hartvigsen-etal-2022-toxigen} & 0.25 & 0.79 & 0.38 & 0.44 & 55 \\ \hline
\textbf{ToxiTwitch (ours)} & \underline{0.61} & 0.88 & \underline{0.72} & \underline{0.79} & {60} \\
\end{tabular}}
\end{table}

\section{Discussion}\label{sec:dis}

In this section, we discuss the key implications of our findings based on our three overarching research questions. 

\paragraph{\textbf{RQ1 (LLM Reasoning):}} How well do LLMs with reasoning capabilities perform in zero-shot classification to detect toxicity in Twitch chat communities? We found that LLMs with reasoning capabilities achieved baseline F1 scores ranging from 0.39 to 0.52. In particular, this was over an initial sample of 500 messages per channel, where the Toxigen model labeled the messages as toxic. However, only 7.8\% of these messages for HasanAbi and 23\% for LolTyler1 were actually toxic according to human annotators.  


\paragraph{\textbf{RQ2 (Emote-Enabled Prompting):}}
 Can augmenting prompts with contextual information about emotes enhance LLM reasoning and improve toxicity detection by reducing misclassification?
Our focus was not on exhaustively benchmarking all possible moderation models (e.g., OpenAI API), but rather on demonstrating how emote-aware prompting and hybrid architectures can contribute to the space.  Our results from the agentic perceive-and-reason implementation—which included various methods for injecting emote context into LLM prompts—demonstrated slight improvements in F1 score, even with a small sample of channel-specific emotes included. However, precision remained low, indicating that while contextual emote information helped to some extent, it did not significantly improve overall classification correctness.

\paragraph{\textbf{RQ3 (Hybrid Model Efficacy):}}
 Can we develop a novel hybrid architecture that combines LLM-generated embeddings with lightweight classical machine learning models to outperform state-of-the-art baselines in both accuracy and latency?
We showed that integrating different agentic components (i.e., perceive-reason-act) of ToxiTwitch and leveraging machine learning models trained on labeled data enabled us to construct a low-latency system with high accuracy and precision. This hybrid model effectively mitigated the limitations of LLM-only approaches and demonstrated improved performance across key metrics.

\subsection{Limitation}

We acknowledge that our dataset is limited to two channels and ~1,000 comments. While this scale restricts generalizability, our goal was to provide a proof-of-concept framework rather than a comprehensive benchmark. Future work should validate ToxiTwitch across broader channel diversity and larger datasets.
 Another limitation of our current approach focusing on EGM is the assumption that the learned representations in embedding space that enable us map channel emotes to global emotes remain consistent overtime. Indeed,~\citet{Moosavi2024} showed that both channel and global emotes meaning changes overtime and thus the similarity association between them may evolve too. These changes mean that for EGM approach to work efficiently we require the embedding space to remain up-to-date. In this   study, we did not face this issue as we purposely collected Twitch chat messages that corresponded to the time-frame that the embedding space was created.  However, we discuss approaches to overcome these two limitation next.

\subsection{Future Work}

Future research could explore addressing some of the observed limitations of our work in two threads. Firstly, we believe that there is a great opportunity to expand our annotation process. Currently, majority of works in LLMs rely on small binary annotation that assist the classifiers in binary prediction tasks. We believe future work could focus on focusing on LLMs reasoning by extracting reasoning syntax tree from  LLMs  and integrate Human in the loop to annotate the structural reasoning of the task as opposed to its binary outcome.
Finally, we intend to investigate the use of continual learning techniques to update the model incrementally without retraining from scratch, ensuring responsiveness to platform dynamics while preserving past knowledge. Although a full ablation of ED vs. EGM vs. text-only embeddings would strengthen the analysis, we see our results as an initial demonstration that both strategies provide measurable gains. A deeper per-category and ablation study is a valuable avenue for future extensions.

\section{Conclusion}\label{sec:con}
The findings from ToxiTwitch highlight the complex interplay between technical design and sociocultural context in automated moderation systems. Emote-based communication on Twitch reflects community-specific and evolving visual languages that LLMs are not yet equipped to interpret fairly or transparently. Our results reveal how toxicity detection can fail when meaning is detached from its local social setting, underscoring the need for \textbf{Responsible} \textbf{AI} approaches that integrate community knowledge and cultural nuance into model development. By combining visual and textual reasoning with human-grounded annotations, ToxiTwitch illustrates a hybrid path toward moderation that is context-aware, latency-efficient, and interpretable. We argue that future Responsible AI frameworks for online moderation should prioritize not only accuracy but also semantic {transparency, annotator diversity, and participatory evaluation}, ensuring that AI systems respect the cultural norms and expressive practices of the communities they aim to serve.


\newpage
\bibliographystyle{ACM-Reference-Format}
\bibliography{references.bib}

@article{liu2023pre,
  title={Pre-train, prompt, and predict: A systematic survey of prompting methods in natural language processing},
  author={Liu, Pengfei and Yuan, Weizhe and Fu, Jinlan and Jiang, Zhengbao and Hayashi, Hiroaki and Neubig, Graham},
  journal={ACM Computing Surveys},
  volume={55},
  number={9},
  pages={1--35},
  year={2023},
  publisher={ACM New York, NY}
}

@inproceedings{amirizaniani2024llms,
author = {Amirizaniani, Maryam and Martin, Elias and Sivachenko, Maryna and Mashhadi, Afra and Shah, Chirag},
title = {Can LLMs Reason Like Humans? Assessing Theory of Mind Reasoning in LLMs for Open-Ended Questions},
year = {2024},
isbn = {9798400704369},
publisher = {Association for Computing Machinery},
address = {New York, NY, USA},
url = {https://doi.org/10.1145/3627673.3679832},
doi = {10.1145/3627673.3679832},
booktitle = {Proceedings of the 33rd ACM International Conference on Information and Knowledge Management},
pages = {34–44},
numpages = {11},
location = {Boise, ID, USA},
series = {CIKM '24}
}

@article{kojima2022large,
  title={Large language models are zero-shot reasoners},
  author={Kojima, Takeshi and Gu, Shixiang Shane and Reid, Machel and Matsuo, Yutaka and Iwasawa, Yusuke},
  journal={Advances in neural information processing systems},
  volume={35},
  pages={22199--22213},
  year={2022}
}

@article{kim2022understanding,
  title={Understanding and identifying the use of emotes in toxic chat on Twitch},
  author={Kim, Jaeheon and Wohn, Donghee Yvette and Cha, Meeyoung},
  journal={Online Social Networks and Media},
  volume={27},
  pages={100180},
  year={2022},
  publisher={Elsevier}
}

@inproceedings{dreier2023toxicity,
  title={Toxicity in Twitch Live Stream Chats: Towards Understanding the Impact of Gender, Size of Community and Game Genre},
  author={Dreier, Lukas and Pirker, Johanna},
  booktitle={2023 IEEE Conference on Games (CoG)},
  pages={1--4},
  year={2023},
  organization={IEEE}
}

@article{han2023hate,
  title={Hate raids on twitch: Echoes of the past, new modalities, and implications for platform governance},
  author={Han, Catherine and Seering, Joseph and Kumar, Deepak and Hancock, Jeffrey T and Durumeric, Zakir},
  journal={Proceedings of the ACM on Human-Computer Interaction},
  volume={7},
  number={CSCW1},
  pages={1--28},
  year={2023},
  publisher={ACM New York, NY, USA}
}

@inproceedings{wohn2019volunteer,
  title={Volunteer moderators in twitch micro communities: How they get involved, the roles they play, and the emotional labor they experience},
  author={Wohn, Donghee Yvette},
  booktitle={Proceedings of the 2019 CHI conference on human factors in computing systems},
  pages={1--13},
  year={2019}
}

@inproceedings{davidson2017automated,
  title={Automated hate speech detection and the problem of offensive language},
  author={Davidson, Thomas and Warmsley, Dana and Macy, Michael and Weber, Ingmar},
  booktitle={Proceedings of the international AAAI conference on web and social media},
  volume={11},
  number={1},
  pages={512--515},
  year={2017}
}

@article{powell2023you,
  title={“You dumb cracker b* tch”: The legitimizing of White supremacy during a Twitch ban of HasanAbi},
  author={Powell, Aisha and Williams-Johnson, Dana},
  journal={New Media \& Society},
  pages={14614448231191776},
  year={2023},
  publisher={SAGE Publications Sage UK: London, England}
}

@inproceedings{steiger2021psychological,
  title={The psychological well-being of content moderators: the emotional labor of commercial moderation and avenues for improving support},
  author={Steiger, Miriah and Bharucha, Timir J and Venkatagiri, Sukrit and Riedl, Martin J and Lease, Matthew},
  booktitle={Proceedings of the 2021 CHI conference on human factors in computing systems},
  pages={1--14},
  year={2021}
}

@article{10.1371/journal.pone.0228723,
    doi = {10.1371/journal.pone.0228723},
    author = {Salminen, Joni AND Sengün, Sercan AND Corporan, Juan AND Jung, Soon-gyo AND Jansen, Bernard J.},
    journal = {PLOS ONE},
    publisher = {Public Library of Science},
    title = {Topic-driven toxicity: Exploring the relationship between online toxicity and news topics},
    year = {2020},
    month = {02},
    volume = {15},
    url = {https://doi.org/10.1371/journal.pone.0228723},
    pages = {1-24},
    number = {2},

}

@article{hellopartner2022toxic,
  title = {The Dangerous Downfall of a Toxic Twitch Audience},
  author = {{Hello Partner}},
  year = {2022},
  month = {November},
  url = {https://hellopartner.com/2022/11/08/the-dangerous-downfall-of-a-toxic-twitch-audience/},
  note = {Accessed: 2024-10-01}
}

@inproceedings{luong-etal-2024-realistic,
    title = "Realistic Evaluation of Toxicity in Large Language Models",
    author = "Luong, Tinh  and
      Le, Thanh-Thien  and
      Ngo, Linh  and
      Nguyen, Thien",
    editor = "Ku, Lun-Wei  and
      Martins, Andre  and
      Srikumar, Vivek",
    booktitle = "Findings of the Association for Computational Linguistics ACL 2024",
    month = aug,
    year = "2024",
    address = "Bangkok, Thailand and virtual meeting",
    publisher = "Association for Computational Linguistics",
    url = "https://aclanthology.org/2024.findings-acl.61",
    doi = "10.18653/v1/2024.findings-acl.61",
    pages = "1038--1047",
}

@article{davange2024toxic,
  title = {Toxic Chat Detection using Deep Learning},
  author = {Davange, Pratik and Chaudhari, Pratik and Patil, ST and Bhojawala, Arwa},
  journal = {International Journal of Engineering Research in Computer Science and Engineering (IJERCSE)},
  volume = {11},
  number = {1},
  pages = {12},
  year = {2024},
  month = {January},
  publisher = {Vishwakarma Institute of Technology (VIT), Pune, Maharashtra, India}
}

@misc{unitaryai_detoxify,
  author = {{Unitary AI}},
  title = {Detoxify},
  howpublished = {\url{https://github.com/unitaryai/detoxify}},
  year = {2024}
}

@misc{nguyen2023finetuningllama2large,
      title={Fine-Tuning Llama 2 Large Language Models for Detecting Online Sexual Predatory Chats and Abusive Texts}, 
      author={Thanh Thi Nguyen and Campbell Wilson and Janis Dalins},
      year={2023},
      eprint={2308.14683},
      archivePrefix={arXiv},
      primaryClass={cs.CL},
      url={https://arxiv.org/abs/2308.14683}, 
}

@inproceedings{10.5555/3600270.3602070,
author = {Wei, Jason and Wang, Xuezhi and Schuurmans, Dale and Bosma, Maarten and Ichter, Brian and Xia, Fei and Chi, Ed H. and Le, Quoc V. and Zhou, Denny},
title = {Chain-of-thought prompting elicits reasoning in large language models},
year = {2024},
isbn = {9781713871088},
publisher = {Curran Associates Inc.},
address = {Red Hook, NY, USA},
booktitle = {Proceedings of the 36th International Conference on Neural Information Processing Systems},
articleno = {1800},
numpages = {14},
location = {New Orleans, LA, USA},
series = {NIPS '22}
}

@inproceedings{nobata2016abusive,
  title={Abusive language detection in online user content},
  author={Nobata, Chikashi and Tetreault, Joel and Thomas, Achint and Mehdad, Yashar and Chang, Yi},
  booktitle={Proceedings of the 25th international conference on world wide web},
  pages={145--153},
  year={2016}
}

@article{Touvron2023LLaMAOA,
  title={LLaMA: Open and Efficient Foundation Language Models},
  author={Hugo Touvron and Thibaut Lavril and Gautier Izacard and Xavier Martinet and Marie-Anne Lachaux and Timoth{\'e}e Lacroix and Baptiste Rozi{\`e}re and Naman Goyal and Eric Hambro and Faisal Azhar and Aurelien Rodriguez and Armand Joulin and Edouard Grave and Guillaume Lample},
  journal={ArXiv},
  year={2023},
  volume={abs/2302.13971},
  url={https://api.semanticscholar.org/CorpusID:257219404}
}

@misc{hasanabi,
  author    = {Twitch},
  title     = {Hasanabi},
  year      = {2024},
  howpublished = {\url{https://www.twitch.tv/hasanabi}},
  note      = {Accessed: 2024-10-07}
}

@inproceedings{Ho2022LargeLM,
  title={Large Language Models Are Reasoning Teachers},
  author={Namgyu Ho and Laura Schmid and Se-Young Yun},
  booktitle={Annual Meeting of the Association for Computational Linguistics},
  year={2022},
  url={https://api.semanticscholar.org/CorpusID:254877399}
}

@inproceedings{Wang2022TowardsUC,
  title={Towards Understanding Chain-of-Thought Prompting: An Empirical Study of What Matters},
  author={Boshi Wang and Sewon Min and Xiang Deng and Jiaming Shen and You Wu and Luke Zettlemoyer and Huan Sun},
  booktitle={Annual Meeting of the Association for Computational Linguistics},
  year={2022},
  url={https://api.semanticscholar.org/CorpusID:254877569}
}

@inproceedings{Moosavi2024,
  author = {Korosh Moosavi and Elias Martin and Muhammad Aurangzeb Ahmad and Afra Mashhadi},
  title = {E2T2: Emote Embedding for Twitch Toxicity Detection},
  booktitle = {Companion of the 2024 Computer-Supported Cooperative Work and Social Computing (CSCW Companion '24)},
  year = {2024},
  location = {San Jose, Costa Rica},
  publisher = {ACM},
  address = {New York, NY, USA},
  pages = {6},
  doi = {10.1145/3678884.3681840},
  isbn = {979-8-4007-1114-5},
  url = {https://doi.org/10.1145/3678884.3681840}
}

@misc{jigsaw-toxic-comment-classification-challenge,
    author = {cjadams and Jeffrey Sorensen and Julia Elliott and Lucas Dixon and Mark McDonald and nithum and Will Cukierski},
    title = {Toxic Comment Classification Challenge},
    year = {2017},
    howpublished = {\url{https://kaggle.com/competitions/jigsaw-toxic-comment-classification-challenge}},
    note = {Kaggle}
}

@misc{hateSonar,
  author = {Davidson, Thomas and Warmsley, Dana and Macy, Michael and Weber, Ingmar},
  title = {Automated Hate Speech Detection and the Problem of Offensive Language},
  year = {2017},
  note = {\url{https://github.com/nikbearbrown/HateSonar}}
}

@inproceedings{hartvigsen-etal-2022-toxigen,
  title = "{ToxiGen}: A Large-Scale Machine-Generated Dataset for Adversarial and Implicit Hate Speech Detection",
  author = "Hartvigsen, Tom and {et al.}",
  booktitle = "Proceedings of the 60th Annual Meeting of the Association for Computational Linguistics (ACL)",
  year = "2022"
}

@article{10.1145/3567568,
author = {Seering, Joseph and Kairam, Sanjay R.},
title = {Who Moderates on Twitch and What Do They Do? Quantifying Practices in Community Moderation on Twitch},
year = {2022},
issue_date = {January 2023},
publisher = {Association for Computing Machinery},
address = {New York, NY, USA},
volume = {7},
number = {GROUP},
url = {https://doi.org/10.1145/3567568},
doi = {10.1145/3567568},
journal = {Proc. ACM Hum.-Comput. Interact.},
month = dec,
articleno = {18},
numpages = {18}
}

@inproceedings{10.1145/3290605.3300390,
author = {Wohn, Donghee Yvette},
title = {Volunteer Moderators in Twitch Micro Communities: How They Get Involved, the Roles They Play, and the Emotional Labor They Experience},
year = {2019},
isbn = {9781450359702},
publisher = {Association for Computing Machinery},
address = {New York, NY, USA},
url = {https://doi.org/10.1145/3290605.3300390},
doi = {10.1145/3290605.3300390},
booktitle = {Proceedings of the 2019 CHI Conference on Human Factors in Computing Systems},
pages = {1–13},
numpages = {13},
location = {Glasgow, Scotland Uk},
series = {CHI '19}
}

@misc{deepseekai2025deepseekr1incentivizingreasoningcapability,
      title={DeepSeek-R1: Incentivizing Reasoning Capability in LLMs via Reinforcement Learning},
      author={DeepSeek-AI},
      year={2025},
      eprint={2501.12948},
      archivePrefix={arXiv},
      primaryClass={cs.CL},
      url={https://arxiv.org/abs/2501.12948},
}

@article{breiman2001random,
  title={Random forests},
  author={Breiman, Leo},
  journal={Machine learning},
  volume={45},
  number={1},
  pages={5--32},
  year={2001},
  publisher={Springer}
}

@article{cortes1995support,
  title={Support-vector networks},
  author={Cortes, Corinna and Vapnik, Vladimir},
  journal={Machine learning},
  volume={20},
  number={3},
  pages={273--297},
  year={1995},
  publisher={Springer}
}

@article{Moon2023AnalyzingNV,
  title={Analyzing Norm Violations in Live-Stream Chat},
  author={Jihyung Moon and Dong-Ho Lee and Hyundong Justin Cho and Woojeong Jin and Chan Young Park and Min-Woo Kim and Jonathan May and Jay Pujara and Sungjoon Park},
  journal={ArXiv},
  year={2023},
  volume={abs/2305.10731},
  url={https://api.semanticscholar.org/CorpusID:258762439}
}

@article{10.1145/3359157,
author = {Jiang, Jialun Aaron and Kiene, Charles and Middler, Skyler and Brubaker, Jed R. and Fiesler, Casey},
title = {Moderation Challenges in Voice-based Online Communities on Discord},
year = {2019},
issue_date = {November 2019},
publisher = {Association for Computing Machinery},
address = {New York, NY, USA},
volume = {3},
number = {CSCW},
url = {https://doi.org/10.1145/3359157},
doi = {10.1145/3359157},
journal = {Proc. ACM Hum.-Comput. Interact.},
month = nov,
articleno = {55},
numpages = {23}
}

@article{ali2025evolving,
  title={Evolving Hate Speech Online: An Adaptive Framework for Detection and Mitigation},
  author={Ali, Shiza and Blackburn, Jeremy and Stringhini, Gianluca},
  journal={arXiv preprint arXiv:2502.10921},
  year={2025}
}

@article{hu2024toxicity,
  title={Toxicity Detection for Free},
  author={Hu, Zhanhao and Piet, Julien and Zhao, Geng and Wagner, David},
  journal={Advances in Neural Information Processing Systems},
  year={2024}
}

@article{welbl2021challenges,
  title={Challenges in detoxifying language models},
  author={Welbl, Johannes and Glaese, Amelia and Uesato, Jonathan and others},
  journal={arXiv preprint arXiv:2109.07445},
  year={2021}
}

@article{poyane2018toxic,
  title={Toxic comment detection in online discussions},
  author={Poyane, M and others},
  journal={Deep Learning Indaba},
  year={2018}
}

@article{jain2021token,
  title={Token-level toxicity detection in online conversations},
  author={Jain, Saahil and others},
  journal={Proceedings of the ACM on Human-Computer Interaction},
  volume={5},
  number={CSCW2},
  pages={1--30},
  year={2021}
}

@article{njeh2025llm,
  title={LLM-based toxicity detection: Opportunities and challenges},
  author={Njeh, Ines and others},
  journal={Computational Linguistics},
  year={2025}
}

@article{yang2023toxbuster,
  title={ToxBuster: A hybrid LLM-human approach to toxicity detection},
  author={Yang, Jie and others},
  journal={Proceedings of the ACM on Human-Computer Interaction},
  volume={7},
  number={CSCW1},
  year={2023}
}

@article{aldahoul2024hybrid,
  title={Hybrid human-AI approaches for content moderation},
  author={Aldahoul, Noura and others},
  journal={Nature Machine Intelligence},
  year={2024}
}

@article{zhu2024advances,
  title={Advances in real-time toxicity detection},
  author={Zhu, Hao and others},
  journal={ACM Computing Surveys},
  year={2024}
}

@misc{twitch2023guidelines,
  title = {Twitch Community Guidelines},
  author = {{Twitch Interactive, Inc.}},
  year = {2023},
  url = {https://safety.twitch.tv/s/article/Community-Guidelines?language=en_US},
  note = {Last updated September 2023},
  howpublished = {Twitch Safety Center}
}

@inproceedings{oikawa2022stacking,
  title={A Stacking-based Efficient Method for Toxic Language Detection on Live Streaming Chat},
  author={Oikawa, Yuto and Nakayama, Yuki and Murakami, Koji},
  booktitle={Conference on Empirical Methods in Natural Language Processing},
  year={2022}
}

@inproceedings{gao2020offensive,
  title={Offensive Language Detection on Video Live Streaming Chat},
  author={Gao, Zhiwei and Yada, Shuntaro and Wakamiya, Shoko and Aramaki, Eiji},
  booktitle={International Conference on Computational Linguistics},
  year={2020}
}

@misc{electroiq2024twitchstats,
  title = {Twitch Statistics 2024: Revenue, Users, Viewership},
  howpublished = {\url{https://electroiq.com/stats/twitch-statistics/}},
  year = {2024},
  note = {Accessed: 2025-05-20}
}

@misc{hartvigsen2022toxigenlargescalemachinegenerateddataset,
      title={ToxiGen: A Large-Scale Machine-Generated Dataset for Adversarial and Implicit Hate Speech Detection}, 
      author={Thomas Hartvigsen and Saadia Gabriel and Hamid Palangi and Maarten Sap and Dipankar Ray and Ece Kamar},
      year={2022},
      eprint={2203.09509},
      archivePrefix={arXiv},
      primaryClass={cs.CL},
      url={https://arxiv.org/abs/2203.09509}, 
}

@misc{li2023blip2bootstrappinglanguageimagepretraining,
      title={BLIP-2: Bootstrapping Language-Image Pre-training with Frozen Image Encoders and Large Language Models}, 
      author={Junnan Li and Dongxu Li and Silvio Savarese and Steven Hoi},
      year={2023},
      eprint={2301.12597},
      archivePrefix={arXiv},
      primaryClass={cs.CV},
      url={https://arxiv.org/abs/2301.12597}, 
}

\newpage
\appendix

\section{Random Forest \& SVM Details}
\label{app:classifiers}

Table~\ref{tab:classifierconfig} summarizes the classifier configurations and evaluation protocol. We use repeated stratified cross-validation to ensure robust performance estimates across class imbalances. All random seeds are fixed for reproducibility.

\begin{table}[h!]
\centering
\small
\caption{Classifier configurations and evaluation protocol for ToxiTwitch hybrid model.}
\label{tab:classifierconfig}
\begin{tabular}{cc}
\hline
\textbf{Component} & \textbf{Configuration} \\
\hline
\multicolumn{2}{c}{\textit{Random Forest}} \\
n\_estimators & 100 \\
class\_weight & balanced \\
random\_state & 42 \\
\hline
\multicolumn{2}{c}{\textit{Linear SVM}} \\
class\_weight & balanced \\
max\_iter & 5000 \\
dual & auto \\
random\_state & 42 \\
\hline
\multicolumn{2}{c}{\textit{Evaluation Protocol}} \\
cross-validation & RepeatedStratifiedKFold \\
n\_splits & 5 \\
n\_repeats & 3 \\
total evaluations & 15 \\
random\_state & 42 \\
\hline
\end{tabular}
\end{table}

\end{document}